\documentclass[lettersize,journal]{IEEEtran}
\usepackage{amsmath,amsfonts}
\usepackage{array}
\usepackage[caption=false,font=normalsize,labelfont=sf,textfont=sf]{subfig}
\usepackage{textcomp}
\usepackage{stfloats}
\usepackage{url}
\usepackage{verbatim}
\usepackage{graphicx}
\usepackage{cite}
\usepackage[ruled]{algorithm2e}
\usepackage{multirow}
\usepackage{pifont}
\usepackage{bm}
\usepackage{booktabs}
\usepackage{colortbl}
\definecolor{mypink}{rgb}{.99,.91,.95}
\definecolor{mygray}{gray}{.9}
\begin{document}

\title{Exploring Invariant Representation for Visible-Infrared Person Re-Identification}

\author{Lei Tan, Yukang Zhang, Shengmei Shen, ~\IEEEmembership{Member,~IEEE,}, Yan Wang, Pingyang Dai, Xianming Lin, Yongjian Wu, Rongrong Ji, ~\IEEEmembership{Senior Member,~IEEE}
\thanks{Lei Tan, Pingyang Dai, Xianming Lin and Rongrong Ji are with the Media Analytics and Computing Laboratory, Department of Artificial Intelligence,
School of Informatics, Xiamen University, Xiamen 361005, China (e-mail:
tanlei@stu.xmu.edu.cn; pydai@xmu.edu.cn; linxm@xmu.edu.cn; rrji@xmu.edu.cn).}
\thanks{Yukang Zhang, and Shengmei Shen are with Pensees Pte Ltd., Singapore (e-mail: zhangyk@pensees.ai; jane.shen@pensees.ai).}
\thanks{Yan Wang is with the Pinterest, Seattle, WA 98100 USA. (e-mail: yanw@pinterest.com).}
\thanks{Yongjian Wu is with the Youtu Laboratory,
Tencent Technology (Shanghai) Co. Ltd, Shanghai 361005, China (e-mail:
littlekenwu@tencent.com;).}}
\markboth{Journal of \LaTeX\ Class Files,~Vol.~14, No.~8, August~2021}%
{Shell \MakeLowercase{\textit{et al.}}: A Sample Article Using IEEEtran.cls for IEEE Journals}


\maketitle

\begin{abstract}
Cross-spectral person re-identification, which aims to associate identities to pedestrians across different spectra, faces a main challenge of the modality discrepancy. In this paper, we address the problem from both image-level and feature-level in an end-to-end hybrid learning framework named robust feature mining network (RFM). In particular, we observe that the reflective intensity of the same surface in photos shot in different wavelengths could be transformed using a linear model. Besides, we show the variable linear factor across the different surfaces is the main culprit which initiates the modality discrepancy. We integrate such a reflection observation into an image-level data augmentation by proposing the linear transformation generator (LTG). Moreover, at the feature level, we introduce a cross-center loss to explore a more compact intra-class distribution and modality-aware spatial attention to take advantage of textured regions more efficiently. Experiment results on two standard cross-spectral person re-identification datasets, i.e., RegDB and SYSU-MM01, have demonstrated state-of-the-art performance.
\end{abstract}

\begin{IEEEkeywords}
Person Re-Identification, Cross-Spectral Retrieval, Representation Learning.
\end{IEEEkeywords}

\section{Introduction}
Person re-identification (re-id), aiming to address the problem of matching people over a distributed set of non-overlapping cameras, has attracted intensive attention in the last few years due to its wide applications in surveillance systems \cite{eom2019learning,Zhai2020ad,zheng2019pyramidal,zheng2019joint}. Since silicon-based digital cameras are naturally sensitive to near-infrared (NIR), most cameras provide extra infrared (IR) images instead of the visible (VIS) images under poor illumination conditions for better visual quality. This in practice puts the Re-ID problem in a cross-spectral setting and requires the approaches to properly handle both the intra-class variance and the even more significant 
modality discrepancies between cross-spectral images~\cite{wang2019learning}.

\begin{figure}[t]
\centering
\includegraphics[height=4.6cm,width=8.4cm]{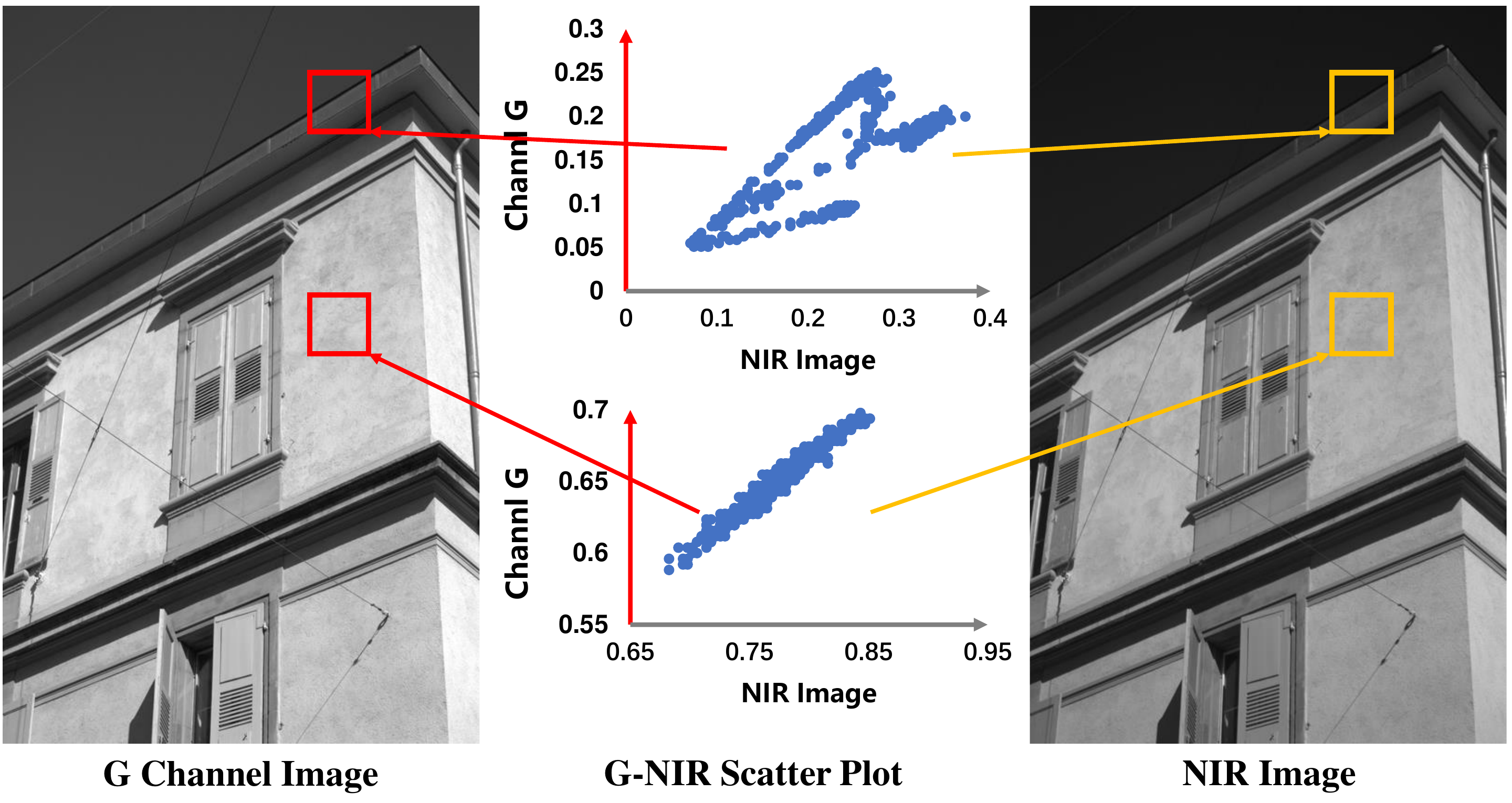}
\caption{\textbf{Illustration of the cross-spectral transformation.} G refers to the green channel of visible image. Under the same illumination, the cross-spectral transformation could be described as a linear transformation in material-similar surface. Still, in the whole image level, the transformation is nonlinear due to the diversity of materials. Since the Re-ID image pairs is not well aligned, we select the cross-spectral image pairs from \cite{brown2011multi}.}
\label{fig:gnir}
\vspace{0em}
\end{figure}

Compared with the single-modality person re-identification problem, the major challenge in such application lies in how to minimize the modality discrepancy between the visible and infrared images. Continuous research efforts have been devoted to addressing this challenge and inspired two typical frameworks. 

One is called feature-based approaches~\cite{dai2018cross,ye2018visible,ye2018hierarchical,lu2020cross,wu2020rgb}, which aim to bridge the apparent gap in the feature space. These approaches typically train a single-path or multi-path Deep Neural Network (DNN) from end to end to obtain a spectrum-invariant embedding. Although the core idea of directly end-to-end learning makes these approaches highly effective and often achieves state-of-the-art performance, this body of works still shows limited ability in minimizing the modality discrepancy. To makes things worse, due to the high complexity and lack of interpretability, the models are hard to adjust or improve.
The other direction is image-level approaches~\cite{wang2019learning, wang2020cross, li2020infrared}, which aim to bridge the appearance discrepancy by transforming the image from one spectrum to another using DNN-based image processing such as Generative Adversarial Networks (GANs)~\cite{goodfellow2014generative}. These approaches usually give better visual verisimilitude and adjustability. However, the lack of a large-scale database providing image pairs across the spectrum makes GAN training challenging and limits the performance of those approaches.

In this paper, we explores the possibility of using actual physics models behind multi-spectral imaging to provide more interpretability, and thus further push the boundary of cross-spectral Re-ID approaches.
Based on the Lambertian reflection model~\cite{horn1986robot,finlayson2005removal}, the illuminations of the same region in VIS and NIR photos should be able to be described using a simple linear model, as long as the region is composed of one consistent material (detail disscussed in Sec. \ref{sec:RP}).
This is illustrated in Figure \ref{fig:gnir}.
Here we use paired VIS-NIR images in the dataset in \cite{brown2011multi}.
For the red and yellow regions in the middle of the image, with a simple linear model, we can accurately predict the pixel values of the NIR image based on the VIS image, as long as the region only has one material. Although the linear transformation exists in the pixel-level of cross-spectral image pairs, the linear factor is determined by the reflection function of the material. It means the linear factor not a constant factor on different surfaces, which causes an image-level non-linear transformation.
In Section \ref{sec:RP}, we analyze and visualize the result to confirm whether the different linear factors in different surfaces are the main culprit that induces the modality discrepancy. It is interesting to find that the modality discrepancy occurs when using variable linear factors among different patches in the image.

The above observation provides us a fresh perspective on the cross-spectral Re-ID task.
Empirically, adopting the observation in the image generation seems the most intuitive way.
As long as we are able to identify regions with different materials and calculate the linear coefficients to transform the input image from one spectrum to another, the modality discrepancy would be easy to bridge.
While this direction may work considering the recent progress on semantic segmentation and a large amount of extra-label for materials, it may be over-complicated and thus not the most effective way.
Even if semantic segmentation works reasonably well, this type of paradigm is still fragile when facing unseen materials.

Driven by this analysis, we extend the aforementioned observation in a feature-level learning framework named robust feature mining (RFM) network to explore a modality-invariant feature representation. Our RFM starts with an easy and efficient linear transformation generator (LTG). To encourage the network to be robust to this kind of linear transformation, the LTG takes effect on randomly selected image patches by a random linear factor and aims to get invariable feature representation to such a transformation. To further mine the invariable feature representation in the feature space, the RFM utilizes the cross-center loss which penalizes the modality discrepancy residual in each part after the single-path DNN model. The cross-center loss employs the modality center instead of the class-center for optimization and shows promising improvement in minimizing the modality discrepancy. Finally, to overcome the tendency of over-focusing on the low modality discrepancy but texture-lacking regions, RFM adopts a modality-aware spatial attention module (MAM) to strengthen the influence of those highly discriminative regions via high-level global features.

We summarize our main contributions as follows:
\begin{itemize}
    \item As an effort to explore the physics model behind the modality discrepancy in the cross-spectral Re-ID task, based on the Lambertian model we demonstrate that the cross-spectral transformation could be described as a linear transformation in regions of the same material and the culprit of modality discrepancy is mainly comes from the image-level non-linear transformation cause by the diversity materials.
    \item By extending the aforementioned diffuse reflection model, we propose a robust feature mining network (RFM). The RFM effectively takes advantage of the above transformation and embeds it in an end-to-end fine-grid framework to explore modality-invariable feature representation.
    \item Extensive experiments on two standard datasets RegDB and SYSU-MM01 prove the effectiveness and superior performance of the proposed RFM.
\end{itemize}

\section{Related Works}
\textbf{Image-level approaches} decompose the modality discrepancy and appearance discrepancies aiming at exploring an efficiency transformation between VIS modality and NIR modality as an image preprocessing. Under this condition, the cross-modality discrepancy is considered as an individual problem alongside the Re-ID problem.
D$^2$RL \cite{wang2019learning} makes the first attempt by using variational autoencoders (VAE) for style disentanglement and generates synthesis images from one domain to another.
Wang \emph{et al}. \cite{wang2020cross} disentangle the features in the encoder and decode from the exchanged features to generate high-quality paired images.
AlignGan \cite{wang2019aligngan} employs a unified GAN framework together with efficient constraints for both image and feature alignment, achieving significant improvement on the quality of synthesized images and final Re-ID performance.
Most recently, X-modality \cite{li2020infrared} designs a light-weight network to learn an intermediate mediator between visible and infrared images instead of focusing on \emph{m}APping visible images to infrared images directly.
Similarly, Ye \emph{et al}. \cite{ye2020visible} utilize generated grayscale images as an assistant for training a more suitable network.
Hi-CMD \cite{choi2020hi} suggests an efficient generator structure that is possible to change the pose and illumination attributes while maintaining the identity characteristic of a specific person.
Although playing a min-max game between the generator and discriminator offers visually impressive results, the generated images are still far from photorealistic, and in turn, limit the final performance.

\textbf{Feature-level approaches} aim at learning modality-invariant feature representations. Wu \emph{et al}. \cite{wu2017rgb} build the largest VIS-NIR Re-ID dataset SYSU-MM01 and propose a zero-padding framework for shared feature learning.
TONE + HCML \cite{ye2018hierarchical} improves the two-stream cross-modality Re-ID framework by jointly optimizing the modality-specific and modality-shared metrics.
Inspired by adversarial learning, cmGan \cite{dai2018cross} adopts a generative adversarial training method to decrease the feature representation discrepancy for different modality images.
Ye \emph{et al}. \cite{ye2020dynamic} propose an attention-based framework to mine the part-level and graph-level contextual cues. Wu \emph{et al}. \cite{wu2020rgb} propose a modality-gated extractor attached with focal modality-aware similarity preserving loss.
Lu \emph{et al}. \cite{lu2020cross} try to introduce modality-specific features based on the cross-modality near the neighbor affinity model.
DG-VAE\cite{pu2020dual} leverages the Gaussian based variational auto-encoder to disentangle an identity-discriminable and an identity-ambiguous cross-modality feature subspace.
However, most of those methods focus on feature representations on top of the entire image. MPANet \cite{wu2021discover} joint modality and pattern alignment network to discover cross-modality nuances for visible-infrared person Re-ID. Due to the diversity of materials, it is difficult to learn an efficient representation in the whole image-level to reduce the modality discrepancy.

\begin{figure*}[t]
\centering
\includegraphics[height=7.8cm,width=17.6cm]{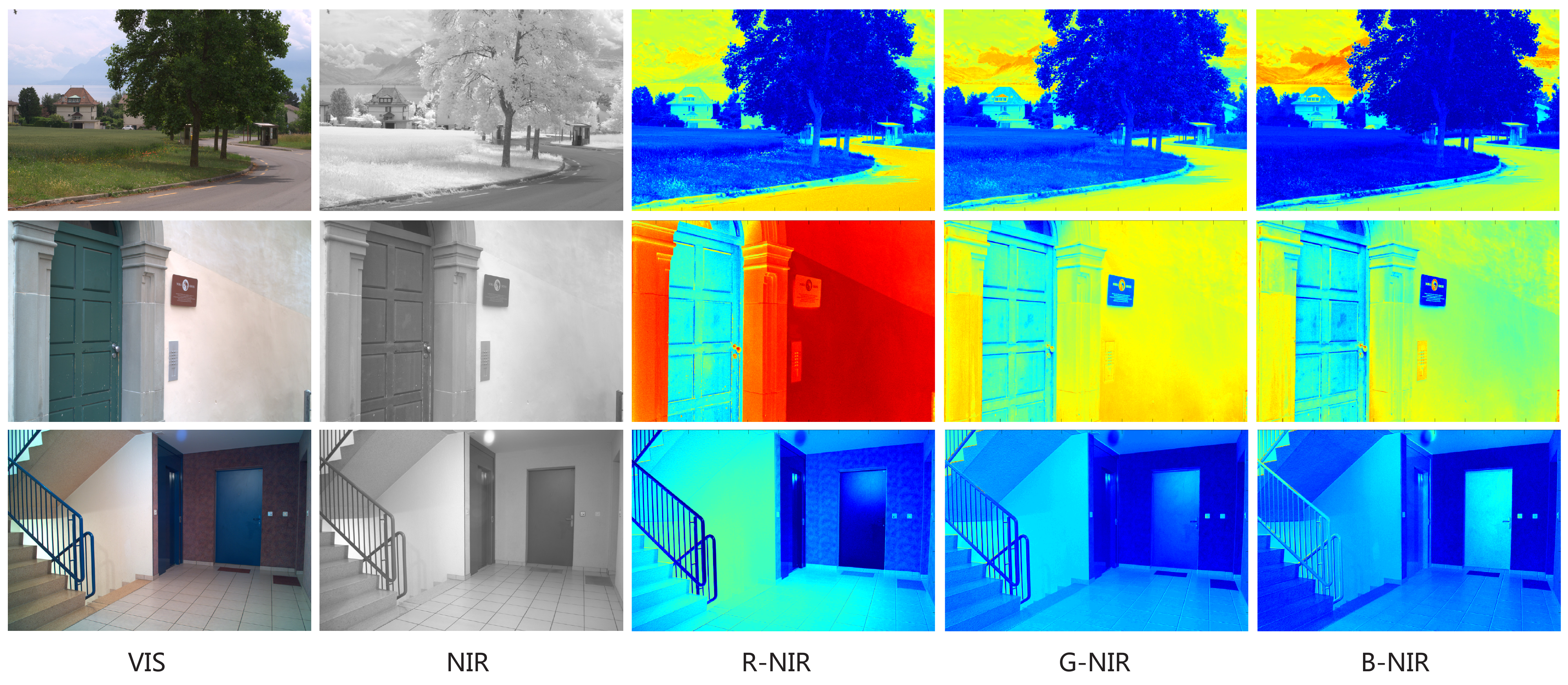}
\caption{\textbf{Example images from the VIS-NIR scene dataset \cite{brown2011multi}}. After we divide the visible image into the red, green, and blue channels and form chromaticity band-ratios from these three spectra and the NIR image, it is clear that the ratio for pixels from the surface with high material-similarity is nearly constant.}
\label{fig:vn}
\end{figure*}

\begin{figure*}[t]
\centering
\includegraphics[height=2.9cm,width=16.8cm]{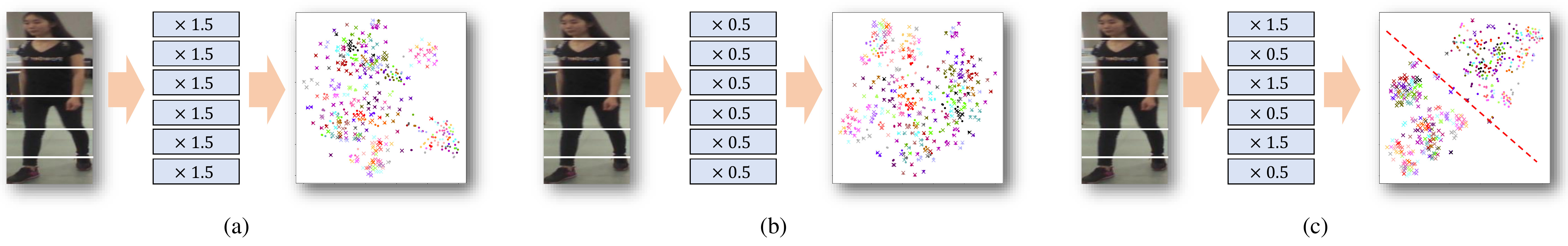}
\caption{\textbf{A example about how modality discrepancy occurs.} Feature space visualization of 100 randomly selected images with/without the local linear transformation. \textbf{(a)$\sim$(b)}: A same linear factor take effect on the whole image will bring limited modality discrepancy. \textbf{(c)}: Variable linear factors take effect on different part shows a huge modality discrepancy. The ’cross’ and ’dot’ marks indicate the samples from the original one and the generated one respectively.}
\label{fig:example1}
\end{figure*}

\section{Reflection Prior for Cross-Spectral Images}
\label{sec:RP}
VIS-NIR matching is a longstanding computer vision problem that has been explored for decades \cite{he2016deep, peng2016graphical}.
One of the main challenges is to formulate and thus alleviate the modality discrepancy.
Using the Lambertian model to analyze the digital image from the multi-sensor cameras is widely applied in some pioneer works\cite{finlayson2005removal, chen2009learning}.
With a light source emitting photons across different wavelengths $\lambda$, the response of each pixel $(x,y)$ in the camera sensors can be formulated as:
\begin{equation}
\label{eq:lamb}
{\rho_j} (x,y) = \sigma (x,y)\int_{\lambda_j}  {{E_j}(\lambda ,x,y)} S(\lambda ,x,y){Q_j}(\lambda )d\lambda,
\end{equation}
where $\lambda$ is the wavelength, $E(\lambda)$ and $S(\lambda)$ denote the spectral power distribution (SPD) of incident light and surface spectral reflectance, and ${Q}(\lambda)$ is the spectral sensitivity of the camera sensor. $j = \left\{ {R,G,B,N} \right\}$ indicates the channel (spectrum). $\sigma (x,y)$ is the Lambertian reflection term which an constant factor and can be calculate by the dot product of the surface normal with the illumination direction.

Following Eq. \ref{eq:lamb}, we leverage a mild assumption to derive a representation between the SPD of the light source and incident light. Generally, we could describe the SPD of light source by a relative spectral power distribution $F(\lambda ,x,y)$ together with a variable $\omega$ which reflects the illumination intensity. We assume that the SPD of incident light in the whole image keeps the same relative spectral power distribution as the light source. Then we could formulate the $E(\lambda ,x,y)$ as:
\begin{equation}
\label{eq:ls}
{E_j}(\lambda ,x,y) = {\beta _j} (x,y){\omega _j}{F_j}(\lambda),
\end{equation}
where the $\beta$ is a parameter to reflect the ratio of intensity between the incident light and the light source.
Then from Eq. \ref{eq:lamb} and Eq. \ref{eq:ls}, if we now consider the images under different spectra like G channel images and NIR images, it is clear that the transformation of G-NIR could be described as:
\begin{equation}
\frac{{{\rho _N}(x,y)}}{{{\rho _G}(x,y)}} = \frac{{{\omega _N}{\beta _N}(x,y)\int_{{\lambda _N}} {{F_N}(\lambda )S(\lambda ,x,y){Q_N}(\lambda )} d\lambda }}{{{\omega _G}{\beta _G}(x,y)\int_{{\lambda _G}} {{F_G}(\lambda )S(\lambda ,x,y){Q_G}(\lambda )d\lambda } }}.
\end{equation}
Under an ideal condition, $\beta$ is supposed to be a high-order term determined by the distance between the light source and the surface \cite{phong1975illumination}.
Now, we could utilize this approximation and regard the $\beta$ as a constant under the same light source to get a simplified expression:
\begin{equation}
\label{eq:exp}
\frac{{{\rho _N}(x,y)}}{{{\rho _G}(x,y)}} = \frac{{{\omega _N}M(x,y,N)}}{{{\omega _G}M(x,y,G)}}.
\end{equation}
Since $F(\lambda)$, ${Q}(\lambda)$, and $S(\lambda)$ are three inner functions depending on the SPD of the light source, the sensitivity of the camera sensor, and the reflection function of the surface material, we replace the Riemann integral with a function $M(x,y,j)$.
In addition, the $\frac{\omega _N}{\omega _G}$ could be considered as a constant factor in two determined spectra.
From this representation, one could observe that in those regions of the same material and under the same illumination condition, the cross-spectral transformation is a linear transformation like shown in Figure \ref{fig:gnir}.
And if we extend it to the entire image, the factor is only influenced by the $S(\lambda ,x,y)$ which is determined by the material.

To verify whether the above equation could be used in various real-life scenarios, we used the paired VIS-NIR scene image dataset introduced by \cite{brown2011multi}. In Figure \ref{fig:vn}, we form chromaticity band-ratios between three VIS spectra and NIR spectrum at each pixel and use the color to reflect the ratio.
We found that the ratio is nearly constant within a region with a consistent material, which holds across R, G, B, and NIR spectra.

After observing the above linear transformation, we further explore whether the variable linear factor in different surfaces is the main culprit that induced the modality gap in such an application. Due to the lacking of a material-based segmentation method to generate according samples, we uniformly segment 100 randomly selected images into six parts from the top to the bottom and multiply each part by a linear factor. Then, we send the new images and original images into an ImageNet \cite{deng2009imagenet} pre-trained Resnet-50 \cite{he2016deep}. Although not so well-aligned, benefitting from the body structure prior from the head to toe, we still find that the modality discrepancy occurs when suffering variable linear factors.

\begin{figure*}[t]
\centering
\includegraphics[height=5cm,width=17.5cm]{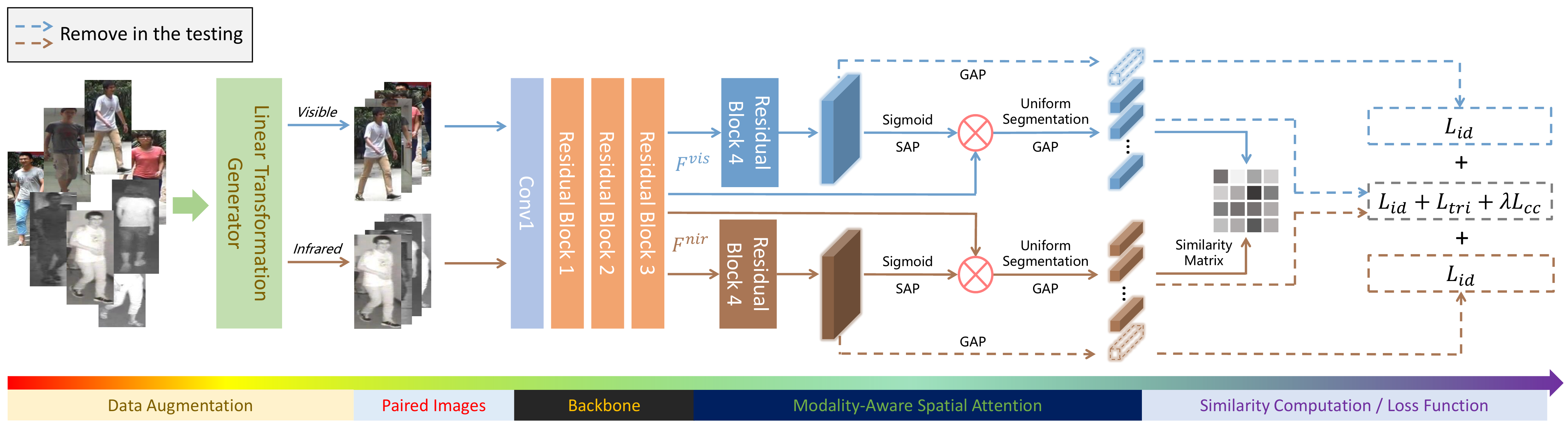}
\caption{\textbf{Framework of the proposed RFM}. \textbf{GAP:} Global Average Pooling. \textbf{SAP:} Spatial-wise Average Pooling. The RFM is built upon a single-path part model with several improvements to adapt to the reflection model.}
\label{fig:framework}
\end{figure*}

\begin{figure*}[t]
\centering
\includegraphics[height=4.9cm,width=16.8cm]{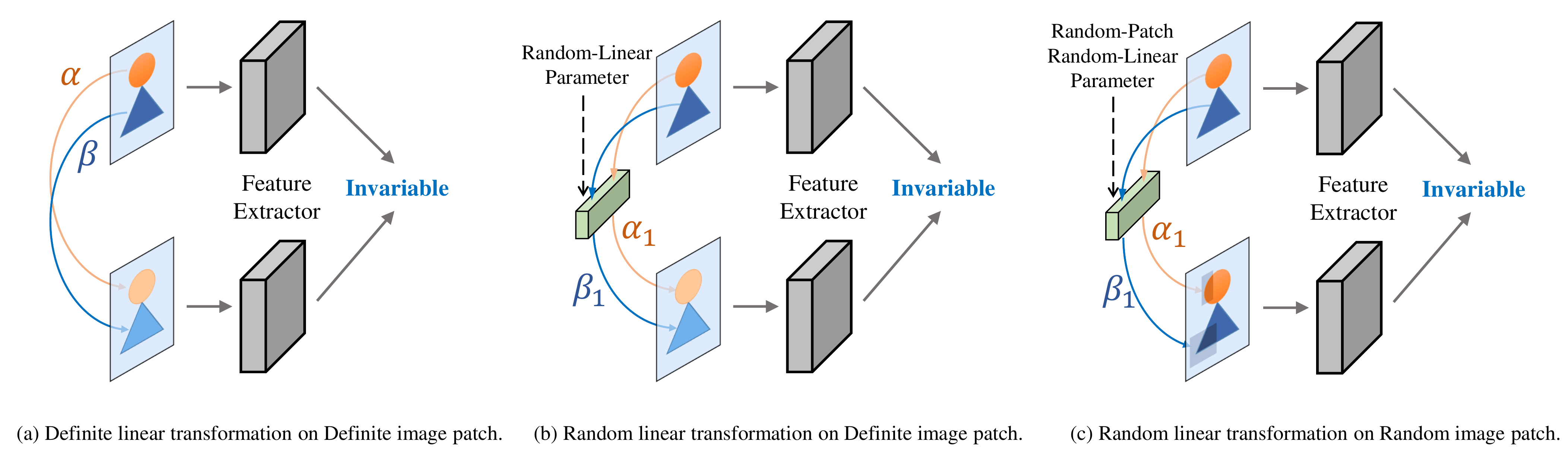}
\caption{\textbf{The motivation of linear transformation generator (LTG).} Herein, we construct an ideal person with only two different surfaces and ignore the background. \textbf{(a)}: As demonstrated above, to obtain a spectral-invariant feature representation, the network should be robust to such a transformation that takes effect upon definite surfaces by definite linear factors. \textbf{(b)}: An ideal data augmentation strategy that takes effect upon definite surfaces by random linear factors. However, this method needs a hard-achieved extra material-aware network for segmentation. \textbf{(c)}: The idea of LTG. By taking effect upon random surfaces by random linear factors, the LTG encourages the network to be robust to a linear transformation anywhere in the image. Under this condition, the cross-spectral transformation can be considered as an easy state of LTG space.}
\label{fig:LTG}
\end{figure*}

\section{Method}
\subsection{Model Architecture}
As illustrated in Figure \ref{fig:framework}, the proposed RFM starts with the linear transformation generator part. Input images $I$ will be transformed to $I^*$. The single-path Resnet-based backbone~\cite{he2016deep} will takes the enhanced images to generate the feature $F$ with spatial dimension $H_{f} \times W_{f}$. 
In this work, we leverage a single-path backbone rather than a multi-path one like in the previous works~\cite{ye2018hierarchical,dai2018cross,lu2020cross,ye2020dynamic} to extract a modality invariant feature embedding.
This structure significantly decreases the computational complexity.
To emphasize those textured regions with high discriminability, we also employ a modality-aware spatial attention module to generate the modality-specific features $F_{att}^v$ and $F_{att}^n$.
Following the PCB model~\cite{sun2018beyond}, feature $F_{att}$ will be uniformly segmented to $K$ parts with spatial dimension $\frac{H_{f}}{K} \times {W_{f}}$ from the height, and then the attached global average pooling layer (GAP) will squeeze all the features to a vector $\bm{V}_{i}$ ($i = 1, 2, \cdots, K$).
Compared with constraining the vectors from each classifier straightly or after the BNNeck~\cite{Luo2019Bags}, we attach the constraints after a fully connected layer to reduce the dimension.
The same strategy is also employed by some previous work \cite{lu2020cross,wang2020cross,wang2019aligngan}.

\begin{algorithm}[t]  
  \caption{linear transformation generator}  
  \label{alg::rle}  
    \KwIn{
      $I$: Input image;
      $C$, $H$ and $W$: Image channel and size;  
      $p$: Probability of the LTG;  
      $s_{min}$ and $s_{max}$: Area of the selected region;
      $r_{min}$ and $r_{max}$: Aspect of the selected region;
      $t_{min}$: Terminate the LTG;}
    \KwOut{Enhanced image $I^{\ast}$}
    \textbf{Initialization:} $p_1\leftarrow$ Rand$ (0, 1)$\;
    \eIf{$p_1 \geq p$}{$I^{\ast}\leftarrow I$\; return $I^{\ast}$\;}
    {$M = Ones(C, H, W)$\; 
    \While{$True$}
    {$S_{r}\leftarrow $Rand$ (s_{min}, s_{max}) \times W \times H$\;       $r_{r}\leftarrow $Rand$ (r_{min}, r_{max})$\;
    $H_{r}\leftarrow \sqrt{S_{r}\times r_{r}}$; $W_{r}\leftarrow \frac{S_{r}}{r_{r}}$\;
    $x_{r}\leftarrow $Rand$ (0, W)$; $y_{r}\leftarrow $Rand$ (0, H)$\;
    \If{$x_{r} + W_{r} \leq W$ and $y_{r} + H_{r} \leq H$}
    {$I_{rle}\leftarrow (C, x_{r}, y_{r}, x_{r} +  W_{r}, y_{r} + H_{r})$\;
    $M_{rle}\leftarrow (C, x_{r}, y_{r}, x_{r} +  W_{r}, y_{r} + H_{r})$\;
    \For{$i\leftarrow 0$ \KwTo $C$}
    {$I_{c}\leftarrow (i, W_{r}, H_{r})$\;
    $M_{c}\leftarrow (i, W_{r}, H_{r})$\;
    $\alpha_{max}\leftarrow \frac{1}{max(I_{c})}$\;
    $\alpha\leftarrow \alpha_{max} \times f_{g}(\cdot)$\;
    $I(I_{c})\leftarrow \alpha \times I_{c}$\;
    $M(M_{c})\leftarrow \alpha \times M_{c}$\;}}
    \If{$min(M) \leq t_{min}$}{$I^{\ast}\leftarrow I$\;
    \textbf{Return} $I^{\ast}$}}}
\end{algorithm}  

\subsection{linear transformation generator}
\label{sec:LTG}
Based on the aforementioned observation, to obtain a modality invariant feature representation, it is necessary to encourage the network to overcome such a transformation that takes effect on a determined surface with a determined linear factor as depicted in Fig. \ref{fig:LTG} (a). Although utilizing random linear factors in different surfaces as shown in Fig. \ref{fig:LTG} (b) seems like the most intuitive way for data augmentation, it highly relies on the well-trained material-aware segmentation network which is hard to achieve. Therefore, we propose the LTG as shown in Fig. \ref{fig:LTG} (c). The LTG randomly selects several image patches and multiplies them with a variable linear parameter. Under the LTG, the cross-spectral transformation could be considered as an easy state of the state space.

\begin{figure}[t]
\centering
\includegraphics[height=5.2cm,width=6.2cm]{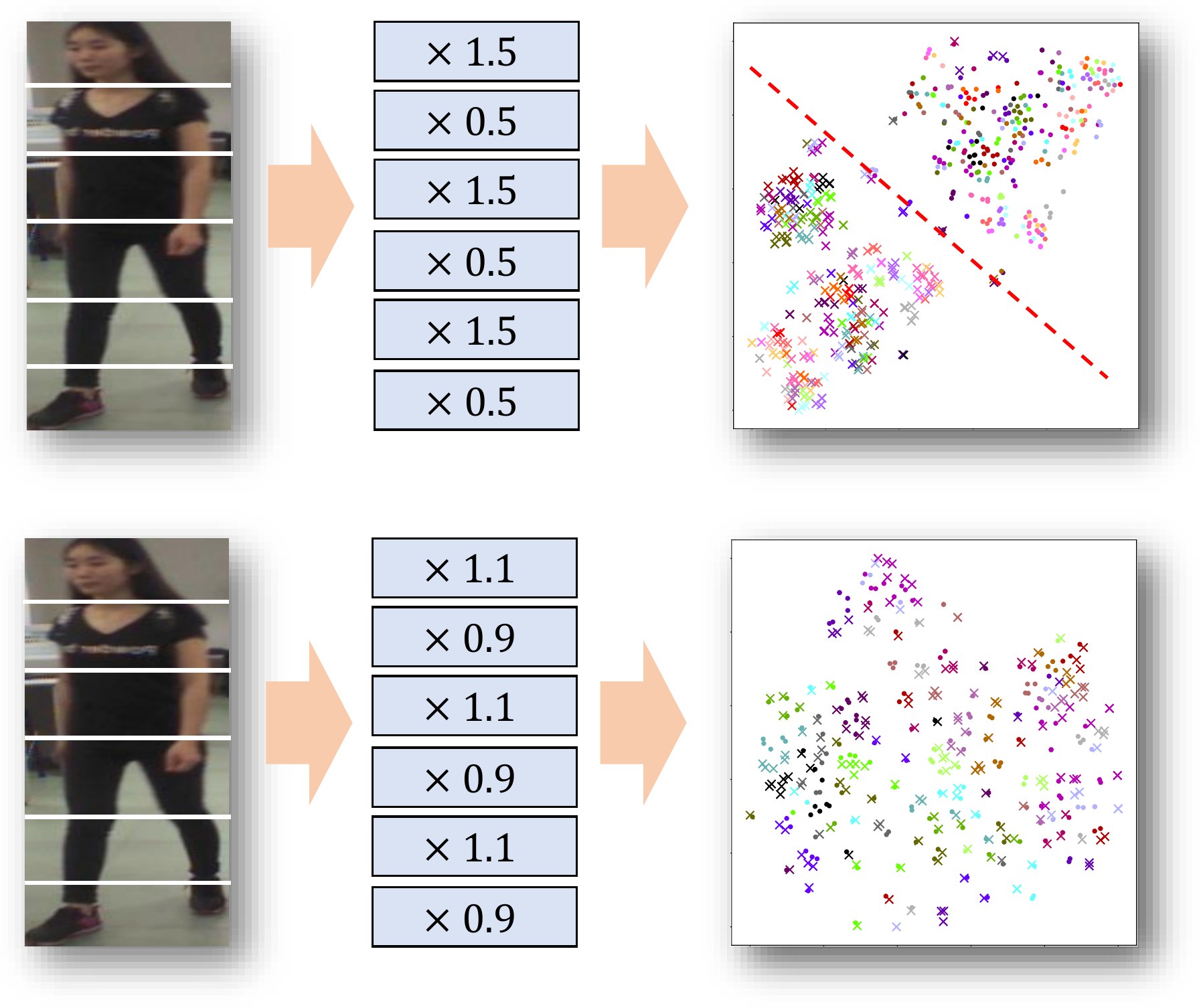}
\caption{\textbf{A example of modality discrepancy.} Small linear factors may not so efficient to generate images with modality gap in the training stage..}
\label{fig:rle}
\end{figure}

Concretely, the LTG is conducting with a certain probability $p$. For an input image $I$, the LTG randomly selects a rectangle region $I_{rle}$ and multiplies it with a linear factor $\alpha$. In case it may exceed the upper bound, we calculate the maximum feasible linear factor in the $I_{rle}$ as $\alpha_{max}$. The linear factor $\alpha$ is calculated by the multiply of $\alpha_{max}$ and a random generator $f_{g}(\cdot)$ on 0 to 1. As shown in Fig. \ref{fig:rle}, a small linear factor can not provide enough modality gap in the training phase. Therefore, a U-shaped Beta distribution is utilized for the $f_{g}(\cdot)$ to obtain high-quality training samples. Furthermore, we set the $s_{min}$ and $s_{max}$ to control area of the selected region and the $r_{min}$ and $r_{max}$ to control the aspect ratio. Then, the region will start on a random selected point $(x_{r}, y_{r})$. The whole processing will be repeated if the rectangle region $I_{rle}$ is out of the boundary. 

Different from the data augmentation strategy like random erasing \cite{zhong2020random} or color-jitter which takes effect on the image only once, the LTG will be repeated several times to obtain a higher modality discrepancy. Therefore, to avoid the vanishing feature, we set a memory matrix $M$ to store the cumulative changes at each pixel. And we set a $t_{min}$ to terminate the LTG when $M(c, x, y) < t_{min}$. Specifically, we show the procedure in Alg. \ref{alg::rle}

\begin{figure}[t]
\centering
\includegraphics[height=5.2cm,width=7.8cm]{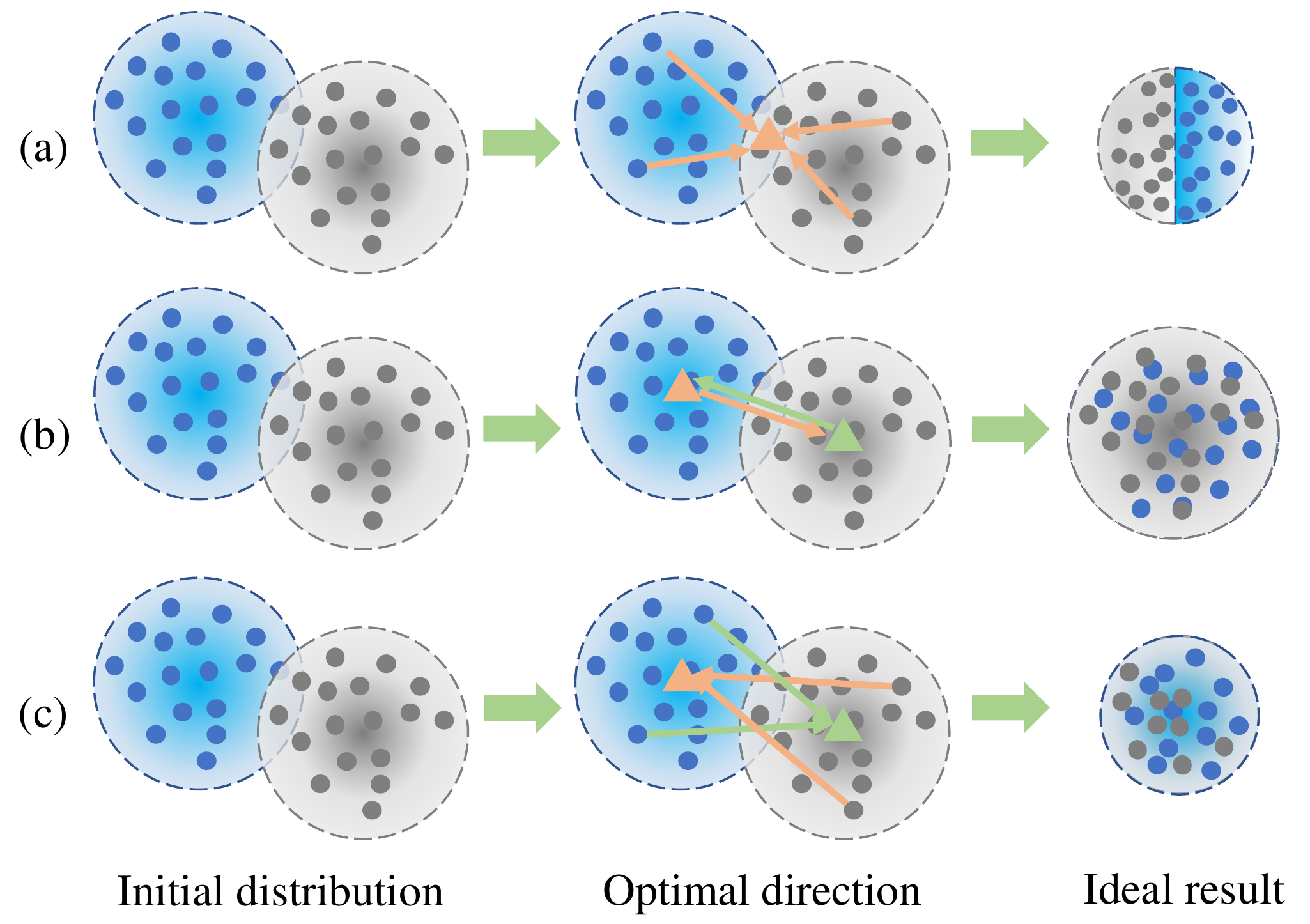}
\caption{\textbf{Comparisons between center loss (a), hetero-center loss (b) and cross-center loss (c).} By exchanging the center of each modality to cluster the samples, the cross-center loss could obtain a more compact distribution.}
\label{fig:ccloss}
\end{figure}

\subsection{Cross-center Loss}
\textbf{Motivation.} Adopting a center-based loss function to directly optimize the modality gap is an intuitive strategy.
The widely used center loss \cite{wen2016discriminative} aims to minimize the intra-class discrepancy by the class center. As illustrated in Figure \ref{fig:ccloss} (a), in center loss, the modality discrepancy is largely ignored which limited the optimization for the modality discrepancy.
Recently, HC-Loss \cite{zhu2020hetero} is proposed to optimize the modality gap by aligning the center of two modalities. However, the HC-loss only focus on the average of the cross-spectral distribution as illustrated in Figure \ref{fig:ccloss} (b), which shows limited performance to compact the final distribution.
To fully utilize the most consistent feature, we introduce a cross-center loss to simultaneously optimize the average and the variance of the cross-spectral distribution as shown in Figure \ref{fig:ccloss} (c).\\

\textbf{Definition.} The cross-center loss is designed to address the modality discrepancy in the cross-spectral Re-ID issue.
As a brief recap, for each input $I_i$, the basic center loss can be formulated as:
\begin{equation}
{\mathcal{L}}_c = \frac{1}{{2}}\sum\limits_{i = 1}^N {D(f({I_i}),{c_{{y_i}}})},
\end{equation}
where $c_{y_i} \in {\mathbb{R}}^d$ refers to the center of class $y_i$ and $d$ denotes the dimension of feature. $f(\cdot)$ refers to the backbone network and $D( \cdot )$ is the distance function, which is Euclidean distance here. $N$ is the size of the mini-batch.
In order to add modality discrepancy into consideration, we divide the $I_i$ into VIS modality $I_i^{v}$ and NIR modality $I_i^{n}$, then use the center of each modality in single classes instead of the original center for optimization.
To simultaneously minimize the intra-class discrepancy as well as the modality discrepancy, we further revise the loss by exchanging the position of two centers:
\begin{equation}
\label{eq:losscc}
{\mathcal{L}}_{cc} = \frac{1}{2}(\sum\limits_{i = 1}^{{N_v}} {D(f(I_i^v),{c_{y_i^n}})}  + \sum\limits_{i = 1}^{{N_n}} {D(f(I_i^n),{c_{y_i^v}})} ).
\end{equation}
Here the $c_{y_{i}^n}$ and $c_{y_{i}^v}$ refer to the two modality centers of class $y_i$. $N_v$ and $N_n$ denote the number of VIS and NIR samples respectively in each mini-batch.

\subsection{Modality-Aware Spatial Attention}
Although using the Cross-Center loss highly improves the performance, the network will easily pay much attention to those textureless regions. These parts provide lower discrepancy but limited discriminative features for re-identification.
An intuitive get-around is to concatenate the feature with the higher-level feature, but it will also bring a thorny problem that the higher-level features with better representation ability will dominate the final output.
To solve the modality discrepancy under limited samples, the higher-level features tend to overfit in the training stage which in turn decreased the final performance.
Therefore, we introduce a modality-aware spatial attention module (MAM) that leverages the high-level feature in every single modality to obtain an attention \emph{m}AP to strengthen the low-level feature.
According to the reflection model, considering the condition that the deep features will be influenced by the modality discrepancy, we use a modality-aware network to match the features with modality discrepancy.
Given the output feature of Residual-Block 3 as $F$, we divide it into $F^{v}$ and $F^{n}$ by its modality. The attention \emph{m}AP $F_{att}^v$ and $F_{att}^n$ is obtained from the outputs of Residual-Block 4 attached with spatial-wise average pooing layer (SAP) and the Sigmoid function.
Specifically, the $F_{att}^v$ and $F_{att}^n$ could be described as:
\begin{equation}
F_{att}^m = \sigma ({f_S}({f_m}(F^m))),m = v, n,
\end{equation}
where we use $v$ and $n$ to indicate the VIS and NIR modalities respectively.
$\sigma(\cdot)$ represents the sigmoid function, $f_S(\cdot)$ represents the spatial-wise average pooing, and $f_{m}(\cdot)$ represents the modality-aware residual blocks.

\subsection{Optimization}
Besides the cross-center loss, the identity-based softmax cross-entropy loss $\mathcal{L}_{id}$ and triplet loss $\mathcal{L}_{tri}$ with online hard-mining \cite{schroff2015facenet} are combined to supervise a more discriminative and robust embedding. Therefore, our loss function for each classifier can be described as:
\begin{equation}
\label{eq:all}
{\mathcal{L}_{cls}} = {\mathcal{L}}_{id} + {\mathcal{L}}_{tri} + \lambda {\mathcal{L}}_{cc}.
\end{equation}

In order to balance different parts, we add a parameter $\lambda$ to adjust the weight of cross-center loss. To constrain the high-level feature, we utilize the softmax cross-entropy for each modality as $\mathcal{L}_{id}^{v}$ for VIS and $\mathcal{L}_{id}^{n}$ for NIR. Therefore, the total loss function could be described as:
\begin{equation}
\mathcal{L} = \mathcal{L}_{id}^{v} + \mathcal{L}_{id}^{n} + \sum\limits_{k = 1}^K {\mathcal{L}_{cls}^k},
\end{equation}
where the $K$ refers to the number of parts.\\
\begin{table*}
  \centering
  \renewcommand{\arraystretch}{1.1}
  \caption{\textbf{Performance on RegDB and SYSU-MM01 datasets}. R-1, 10, 20 denotes the Rank-1, 10, 20 accuracy. * denotes the method using multi-query in the testing phase.}
  \resizebox{\textwidth}{!}{
  \begin{tabular}{rccccccccccccccccc}
  \toprule[1pt]
   \multirow{3}{*}{Model} & \multirow{3}{*}{Pub.}  & \multicolumn{8}{c}{RegDB} & \multicolumn{8}{c}{SYSU-MM01} \\
   \cline{3-18}
   & &\multicolumn{4}{c}{Visible to Thermal} &\multicolumn{4}{c}{Thermal to Visible} &\multicolumn{4}{c}{All Search} &\multicolumn{4}{c}{Indoor Search}\\
   \cline{3-18}
   & & R-1 & R-10 & R-20 & \emph{m}AP   & R-1 & R-10 & R-20 & \emph{m}AP & R-1 & R-10 & R-20 & \emph{m}AP & R-1 & R-10 & R-20 & \emph{m}AP\\
   \hline
    Zero-Padding\cite{wu2017rgb}        & ICCV'17   & 17.8 & 34.2 & 44.4 & 18.9 & 16.6 & 34.7 & 44.3 & 17.8 & 14.8 & 54.1 & 71.3 & 15.9 & 20.6 & 68.4 & 85.8 & 26.9 \\
    HCML\cite{ye2018hierarchical}       & AAAI'18   & 24.4 & 47.5 & 56.8 & 20.8 & 21.7 & 45.0 & 55.6 & 22.2 & 14.3 & 53.2 & 69.2 & 16.2 & 24.5 & 73.3 & 86.7 & 30.1 \\
    BDTR\cite{ye2018visible}            & IJCAI'18  & 33.6 & 58.6 & 67.4 & 32.8 & 32.9 & 58.5 & 68.4 & 32.0 & 17.0 & 55.4 & 72.0 & 19.7 & - & - & - & - \\
    cmGAN\cite{dai2018cross}            & IJCAI'18  & -    & -    & -    & -    & -    & -    & -    & -  & 27.0 & 67.5 & 80.6 & 27.8 & 31.7 & 77.2 & 89.2 & 42.2   \\
    HSME\cite{hao2019hsme}              & AAAI'19   & 50.9 & 73.4 & 81.7 & 47.0 & 50.2 & 72.4 & 81.1 & 46.2 & 20.7 & 62.8 & 78.0 & 23.2 & - & - & - & - \\
    AlignGAN\cite{wang2019aligngan}     & ICCV'19   & 57.9 & -    & -    & 53.6 & 56.3 & - & - & 53.4 & 42.4 & 85.0 & 93.7 & 40.7 & 45.9 & 87.6 & 94.4 & 54.3 \\
    X-Modality\cite{li2020infrared}     & AAAI'20   & 62.2 & 83.1 & 91.7 & 60.2 & - & - & - & - & 49.9 & 89.8 & 96.0 & 50.7 & -    & -    & -    & -    \\
    MGE + FMASP\cite{wu2020rgb}         & IJCV'20   &65.1 & 83.7 & - & 64.5 & - & - & - & - & 43.6 & 86.3 & - & 45.0 & 48.6 & 89.5 & - & 57.5 \\
    DG-VAE\cite{pu2020dual}             & MM'20     & 73.0 & 86.9 & - & 71.8 & - & - & - & - & 59.5 & 93.8 & - & 58.5 & - & - & - & - \\
    Hi-CMD\cite{choi2020hi}             & CVPR'20   & 70.9 & 86.4 & - & 66.0 & - & - & - & - & 34.9 & 77.6 & - & 35.9 & - & - & - & - \\
    SSFT\cite{lu2020cross}             & CVPR'20   & 72.3 & -    & - & 72.9 & 71.0 & - & - & 71.7 & 61.6 & 89.2  & 93.9 & 63.3  & 70.5 & 94.9 & 97.7 & 72.6 \\
    DDAG\cite{ye2020dynamic}            & ECCV'20   & 69.3 & 86.2 & 91.5 & 63.5 & 68.1 & 85.2 & 90.3 & 61.8 & 54.8 & 90.4 & 95.8 & 53.0 & 61.0 & 94.1 & 98.4 & 68.0 \\
    CICL + IAMA\cite{zhao2021joint}     & AAAI'21   & 78.8 & -    & - & 69.4 & 77.9 & - & - & 69.4 & 57.2 & 94.3 & 98.4 & 59.3 & 66.6 & 98.8 & 99.7 & 74.7 \\
    MCLNet\cite{hao2021cross}           & ICCV'21   & 80.3 & 92.7 & 96.0 & 73.1 & 75.9 & 90.9 & 94.6 & 69.5 & 65.4 & 93.3 & 97.1 & 62.0 & 72.6 & 97.0 & 99.2 & 76.6 \\
    SMCL\cite{wei2021syncretic}         & ICCV'21   & 83.9 & -    & - & 79.8 & 83.1 & - & - & 78.6 & 67.4 & 92.9 & 96.8 & 61.8 & 68.8 & 96.6 & 98.8 & 75.6 \\
    VCD + VML\cite{tian2021farewell}    & CVPR'21   & 73.2 & -    & - & 71.6 & 71.8 & - & - & 70.1 & 60.0 & 94.2 & 98.1 & 58.8 & 66.1 & 96.6 & 99.4 & 73.0  \\
    NFS\cite{chen2021neural}            & CVPR'21   & 80.5 & 92.0 & 95.1 & 72.1 & 78.0 & 90.5 & 93.6 & 69.8 & 56.9 & 91.3 & 96.5 & 55.5 & 62.8 & 96.5 & 99.1 & 69.8 \\
    MPANet\cite{wu2021discover}         & CVPR'21   & 82.8 & -    & - & 80.7 & 83.7 & - & - & 80.9 & 70.6 & 96.2 & 98.8 & 68.2 & 76.7 & 98.2 & 99.6 & 81.0 \\
    MMN\cite{zhang2021towards}         & MM'21   & 91.6 & 97.7 & 98.9 & 84.1 & 87.5 & 96.0 & 98.1 & 80.5 & 70.6 & 96.2 & 99.0 & 66.9 & 76.2 & 97.2 & 99.3 & 79.6 \\
    \hline
    \rowcolor{mygray}\textbf{RFM}                         & This work & \textbf{93.5} & \textbf{98.4} & \textbf{99.3} & \textbf{87.5} & \textbf{91.0} & \textbf{97.5} & \textbf{98.6} & \textbf{86.6} & \textbf{72.5} & \textbf{97.7} & \textbf{99.6} & \textbf{70.5} & \textbf{81.1} & \textbf{99.6} & \textbf{100.0} & \textbf{84.6} \\
    \toprule[1pt]
    \end{tabular}}
    \label{Tbl:final}
\end{table*}

\section{Experiments}
\subsection{Experimental Setup}
\textbf{Dataset.} The proposed RFM is evaluated on two public available cross-spectral Re-ID datasets RegDB \cite{nguyen2017person} and SYSU-MM01 \cite{wu2017rgb}. The Cumulative Matching Characteristic (CMC) and mean Average Precision (\emph{m}AP) are used as evaluation metrics.

RegDB consists of 412 persons, where each person has 10 RGB images and 10 IR images. We evaluate in both visible-to-infrared and infrared-to-visible settings. Following the evaluation protocol of \cite{ye2018hierarchical}, we repeat 10 trails with a randomly half-half split of the dataset, one-half identities for training and the other for testing. The final results are based on an average of 10 times testing.

SYSU-MM01 is a large-scale dataset collected by four visible cameras and two infrared cameras, including indoor and outdoor environments.
The training set contains $22,258$ RGB images and $11,909$ IR images of $395$ identities, and the testing set contains $96$ identities with $3,803$ IR images as the query.
The gallery set is determined by the testing mode, which includes all-search and indoor-search. In the all-search mode, all of the images from visible cameras are used as the gallery, while in the indoor-search model, only two indoor visible cameras are used as the gallery.

\textbf{Implementation details.} We employ the Resnet-50~\cite{he2016deep} pre-trained on ImageNet \cite{deng2009imagenet} as the backbone network.
We resize all the input images to $384 \times 192$ and adopt commonly used horizontal flipping as data augmentation.
Following \cite{ye2020dynamic}, in each mini-batch, we randomly select 7 identities with 8 visible images and 8 infrared images for training.
SGD is utilized as the optimizer.
During the training stage, the learning rate is initiated from $0.03$ and is decayed to $0.003$ and $0.0003$ respectively after 40th and 70th epoch. 
Following the random erasing \cite{zhong2020random}, the probability of the LTG is set to 0.5,  $s_{min} = 0.02$, $s_{max} = 0.4$, $r_{min}=\frac{1}{r_{max}}=0.3$ and $t_{min}=1e-6$. For the CC-Loss, we set its weight $\lambda$ as 3.

\subsection{Comparison with state-of-the-art}
In Table \ref{Tbl:final}, we evaluate RFM against the previously reported state-of-the-art methods on the RegDB and SYSU-MM01.
For all tracks, we could observe that the proposed RFM outperforms the previous methods, which indicates the great adaptability of RFM in the cross-spectral person re-id task.
Compared with the SYSU-MM01, samples in the RegDB dataset contain fewer occlusion, less pose variation, and more consistent illumination.
As mentioned above, it is much harder to minimize the modality discrepancy in a complex environment.
That may be the reason why RFM achieves a larger improvement in RegDB rather than SYSU-MM01.
Additionally, compared with the previous methods, the proposed RFM is more efficient with better interpretability and could be used together with other popular approaches.

\begin{table}[t]
  \centering
  \renewcommand{\arraystretch}{1.2}
  \caption{\textbf{Ablation study on SYSU-MM01.} LTG, CC, and MAM refers to the linear transformation generator, cross-center loss and modality-aware spatial attention.}
  \resizebox{80mm}{!}{
  \begin{tabular}{lccccccc}
  \toprule[1pt]
   \multirow{2}{*}{\textbf{Model}} & \multicolumn{3}{c}{\textbf{Settings}} & \multicolumn{4}{c}{\textbf{SYSU-MM01}} \\
    & REL & CC & MAM & \emph{R-1} & \emph{R-10} & \emph{R-20} & \emph{\emph{m}AP}\\
   \hline
   \hline
    Baseline    &           &           &           & 60.3 & 93.4 & 97.5 & 58.2\\
    +LTG        & \ding{51} &           &           & 65.5 & 95.5 & 98.4 & 62.0\\
    +LTG+CC     & \ding{51} & \ding{51} &           & 70.9 & 97.2 & 99.3 & 67.9\\
    +LTG+MAM    & \ding{51} &           & \ding{51} & 66.4 & 95.8 & 98.9 & 63.0\\
    \rowcolor{mygray}+LTG+CC+MAM   & \ding{51} & \ding{51} & \ding{51} & \textbf{72.5} & \textbf{97.7} & \textbf{99.6} & \textbf{70.5}\\
   \bottomrule[1pt]
    \end{tabular}}
    \label{Tbl:ablation}
\end{table}

\subsection{Ablation Study}
Here, we evaluate the influence of the proposed architectural components with different settings and show the quantitative results in Table \ref{Tbl:ablation}. In particular, we select the Resnet-50 backbone with 12 parts PCB strategy as the baseline and conduct the whole experiments in the SYSU-MM01 all-search mode. 

Compared to the baseline, add the linear transformation generator could highly improve the Rank-1 accuracy and \emph{m}AP as 5.2\% and 3.8\% in respectively. Further adding the cross-center loss (CC Loss) also obtain a great gain to the model. The modality-aware attention module is designed to overcome the tendency of over-focusing on the low modality discrepancy caused by the use of CC Loss. Therefore, it shows better effectiveness when combines with CC Loss.

\begin{table}[t]
  \centering
  \renewcommand{\arraystretch}{1.2}
  \caption{\textbf{Comparison of the different types of LTG on SYSU-MM01.}}
  \resizebox{80mm}{!}{
  \begin{tabular}{lccccccc}
  \toprule[1pt]
   \multirow{2}{*}{\textbf{Model}} & \multirow{2}{*}{\textbf{Settings ($f_{g}(\cdot)$)}} & \multicolumn{4}{c}{\textbf{SYSU-MM01}} \\
   &    &\emph{R-1} & \emph{R-10} & \emph{R-20} & \emph{\emph{m}AP}\\
   \hline
   \hline
    Random Erasing\cite{zhong2020random}        & Constant & 70.6 & 95.6 & 98.7 & 68.0\\
    LTG                   & Equal weight   & 69.2 & 96.7 & 99.1 & 66.3\\
    LTG                   & Beta(0.3, 0.3) & 72.3 & 97.5 & 99.5 & 70.2\\
    \rowcolor{mygray}LTG  & Beta(0.5, 0.5) & \textbf{72.5} & \textbf{97.7} & \textbf{99.6} & \textbf{70.5}\\
   \bottomrule[1pt]
    \end{tabular}}
    \label{Tbl:rle}
\end{table}

\begin{figure}[t]
\centering
\includegraphics[height=3.0cm,width=7.8cm]{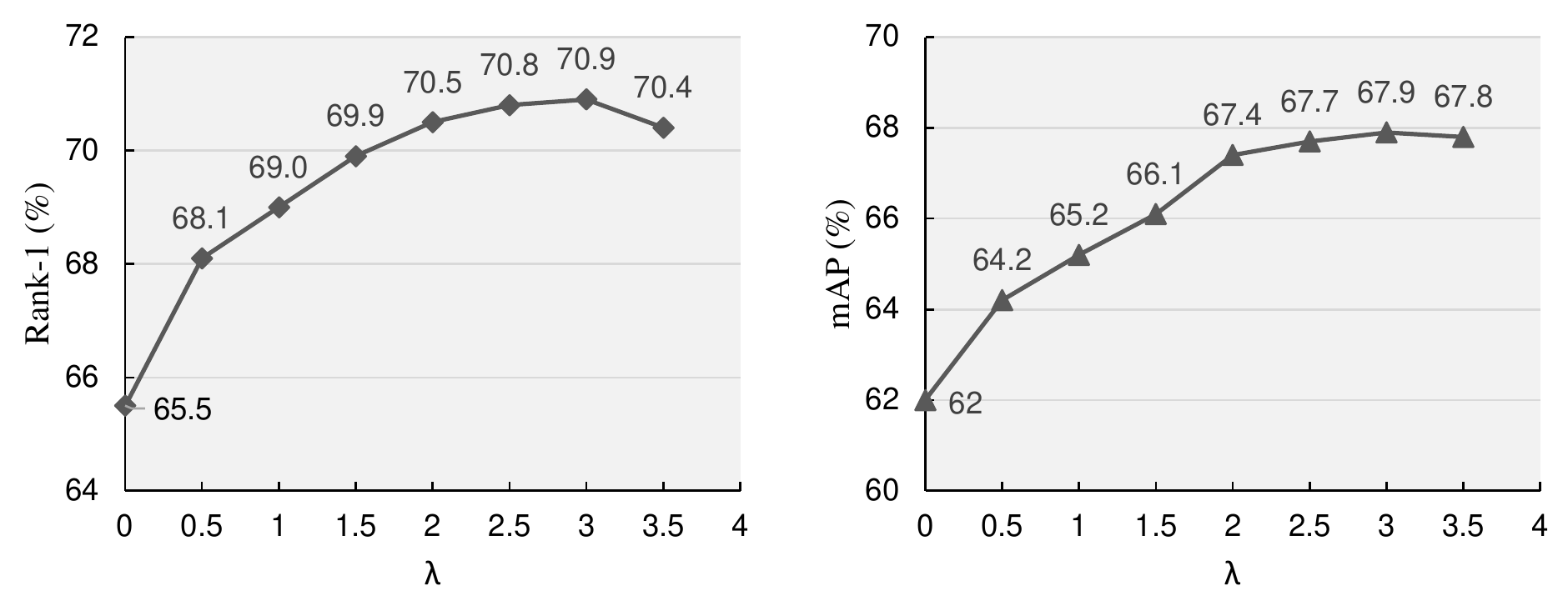}
\caption{\textbf{Validation results of Cross-Center Loss on SYSU-MM01}. Compared with the previous works aforementioned, the CC-Loss shows significant improvements in this task.}
\label{fig:cc}
\end{figure}

\begin{table}[t]
  \centering
  \renewcommand{\arraystretch}{1.2}
  \caption{\textbf{Comparison of the different types of spatial attention strategy on SYSU-MM01.} Since the scale of HC-Loss hasn't been normalized by the batch size, for fair comparison, we test both default and re-implement settings.}
  \resizebox{80mm}{!}{
  \begin{tabular}{lcccccc}
  \toprule[1pt]
   \multirow{2}{*}{\textbf{Model}} & \multirow{2}{*}{\textbf{Settings}}& \multicolumn{4}{c}{\textbf{SYSU-MM01}} \\
   & &\emph{R-1} & \emph{R-10} & \emph{R-20} & \emph{\emph{m}AP}\\
   \hline
   \hline
    Baseline(+LTG) &Center-Loss(3.0)  & 67.5 & 96.7 & 99.2 & 66.0\\
    Baseline(+LTG) &HC-Loss(0.5) & 68.7 & 96.3 & 98.9 & 65.1\\
    Baseline(+LTG) &HC-Loss*(3.0)& 69.4 & 96.5 & 99.0 & 65.7\\
     \rowcolor{mygray}Baseline(+LTG) &CC-Loss(3.0)      & \textbf{70.9} & \textbf{97.2} & \textbf{99.3} & \textbf{67.9}\\
   \bottomrule[1pt]
    \end{tabular}}
    \label{Tbl:cc}
\end{table}

\begin{table}[t]
  \centering
  \renewcommand{\arraystretch}{1.2}
  \caption{\textbf{Comparison of the different types of spatial attention strategy on SYSU-MM01.} }
  \resizebox{80mm}{!}{
  \begin{tabular}{lcccccc}
  \toprule[1pt]
   \multirow{2}{*}{\textbf{Model}} & \multirow{2}{*}{\textbf{Settings}}& \multicolumn{4}{c}{\textbf{SYSU-MM01}} \\
   & &\emph{R-1} & \emph{R-10} & \emph{R-20} & \emph{\emph{m}AP}\\
   \hline
   \hline
    RFM &w/o attention & 70.9 & 97.2 & 99.3 & 67.9\\
    RFM &Spatial-Attention\cite{woo2018cbam} & 69.6 & 96.1 & 98.8 & 65.1\\
    \rowcolor{mygray}RFM &MAM                                 & \textbf{72.5} & \textbf{97.7} & \textbf{99.6} & \textbf{70.5}\\
   \bottomrule[1pt]
    \end{tabular}}
    \label{Tbl:mam}
\end{table}

\begin{figure}[t]
\centering
\includegraphics[height=4.2cm,width=8cm]{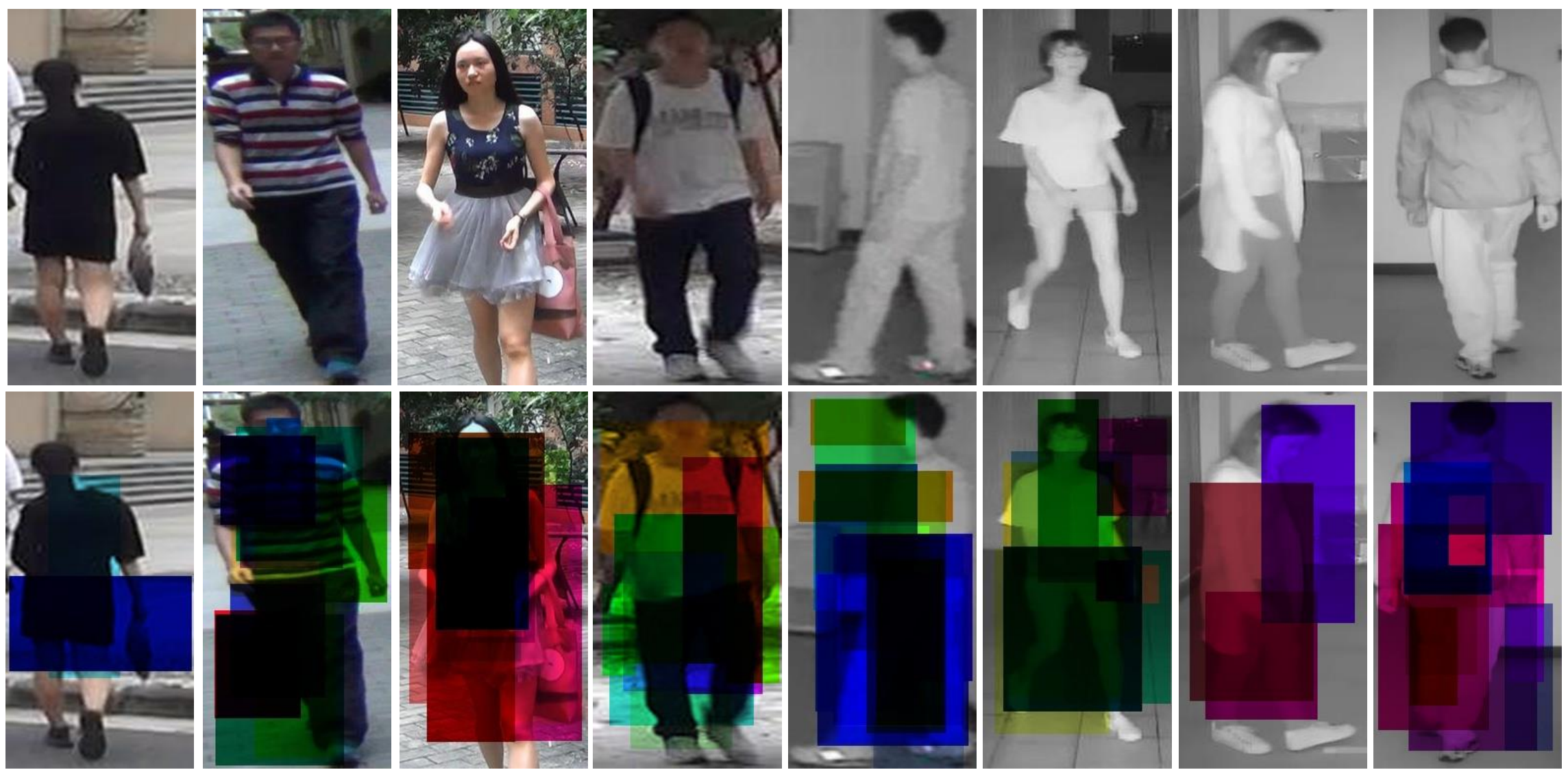}
\caption{\textbf{Visualization results of LTG}. The upper is the input images and the lower is the augmented image. The LTG significantly destroys the modality-relevant information like color while maintaining the modality-irrelevant information like edge.}
\label{fig:aug}
\end{figure}

\subsection{Discussion}
\textbf{LTG.}
The LTG uses the linear function to generate image samples with modality discrepancy. Since we utilize a random generator $f_{g}(\cdot)$ to control the transformation, we make a comparison between different $f_{g}(\cdot)$ to show its performance. As shown in Table \ref{Tbl:rle}, the U-shape beta shape distribution significantly improves the final performance when compared to the uniform distribution. The best performance occurs when the $f_{g}(\cdot)$ is set to Beta(0.5, 0.5). In addition, we also add random erasing into the comparison. Actually, the random erasing could be regarded as a very special case of LTG, when the $f_{g}(\cdot)$ is a constant factor and takes effect only once. Also, the LTG still shows superior performance than the random erasing. We visualize the augmented images in Fig. \ref{fig:aug}. As is known to all, the modality-relevant information mainly comes from the high similarity in colors, which is considered to be eliminated to train a cross-spectral model. Obviously, the LTG-generated images destroy the color similarity in the single spectral sample pairs by a high modality discrepancy, while maintaining the modality-irrelevant information like edge. Therefore, the LTG largely enforces the network to focus on modality-irrelevant information.

\textbf{CC-Loss.}
In Eq. \ref{eq:all}, we set a parameter $\lambda$ to control the trade-off between the cross-center loss with identity loss and triplet loss.
To explore the impact of the hyperparameter, we give an empirical analysis on the SYSU-MM01 datasets and report the results in Figure \ref{fig:cc}. From the results, we can observe that even adding the cross-center loss with a small weight (0.5), the final accuracy and \emph{m}AP could improve significantly. The best performance is achieved when the parameter $\lambda$ goes to 3.
Although a bigger $\lambda$ may obtain a more compact feature space for every single identity, the high weight of cross-center loss makes limited optimization for the feature from different identities. In addition, we compare the CC-Loss with some previous works like HC-Loss \cite{zhu2020hetero} and center loss \cite{wen2016discriminative} to demonstrate the effectiveness of CC-Loss. For a fair comparison, we conduct all the experiments in the LTG-augmented baseline method and show the results in Fig \ref{Tbl:cc}. Besides the default setting as HC-Loss (0.5), we conduct another version as HC-Loss(3.0) for a fair comparison. 
For HC-Loss(3.0), its value is divided by the batch size to keep the same setting with the center loss and the cross-center loss.
Although both of HC-Loss and center loss could provide a positive influence in the cross-modality person re-identification, CC-Loss still shows better adaptability in this task.

\textbf{Modality-aware spatial attention.} 
To demonstrate the effectiveness of the modality-aware spatial attention module (MAM), we compare the modality-aware spatial attention with the widely used spatial attention module \cite{woo2018cbam} and show the results in Table \ref{Tbl:mam}. We observe that straightly adding the spatial attention module in the current layer can not provide more effective feature representation and decrease the final performance. However, MAM performs much better than the original spatial attention module by combining the higher-layer information as a guidance.

\begin{figure}[t]
\centering
\includegraphics[height=4.2cm,width=8cm]{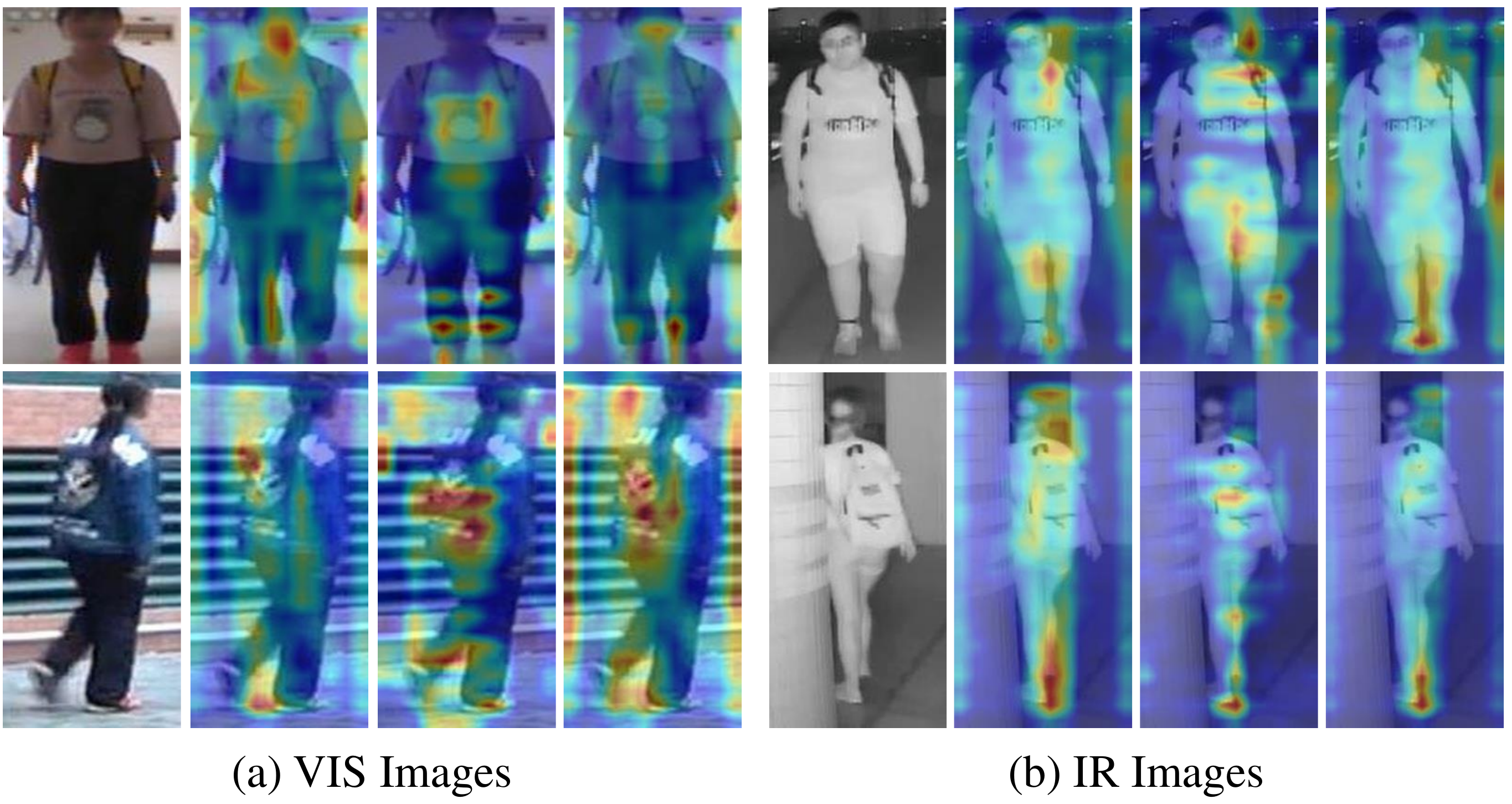}
\caption{\textbf{Visualization results of MAM}. (a) and (b) refers to the VIS images and IR images separately. And from the left to right is the \textbf{input image}, \textbf{original feature map}, \textbf{attention map}, and \textbf{feature \emph{m}AP after the MAM}.  Compared with the original feature \emph{m}AP, MAM helps the features focus more on those discriminative regions like the logos and bag in the above.}
\label{fig:attention}
\end{figure}

\section{Limitations}
In the foundation of the specific observation in the cross-spectral person re-identification, the proposed LTG may not be a universal data augmentation strategy like random flipping or random erasing. Whether breaking the modality-similarity between the image pairs could make sense in other computer vision tasks still need to be evaluated. On the other hand, the scale of the RegDB and SYSU-MM01 datasets are small. Although the proposed RFM reaches state-of-the-art results in both two datasets,  the performance of every component in an open-world scenario hasn't been verification. 
Meanwhile, for the reflection model, we only considered a diffuse refection model, without explicit modeling specularities and higher-order reflections.
In addition, expressing the $E(\lambda)$ by the SPD of the light source includes an assumption that each part gets the same illumination from the light source.
It may not be true for shadows, but does not affect the Re-ID task largely, because the regions in the shadow could be treated as a different material.
Overall, we believe the reflection model, if used wisely, could still provide a new perspective to effectively solve the modality discrepancy in cross-spectral tasks.

\section{Conclusion}
In this paper, we address the modality discrepancy in the cross-modality person re-identification task with the robust feature mining network (RFM). 
We make an attempt to explore how the modality discrepancy occurs in the cross-spectral Re-ID problem and find that the diversity of linear factors across different surfaces is the main culprit. 
By extending the observation, we introduced the linear transformation generator with good interpretability to mine a more robust feature space for this task. Meanwhile, we adopt the fine-grid backbone with the proposed cross-center loss to minimize the modality discrepancy in the parts. 
In addition, a modality-aware spatial attention module is attached to guide the network to pay more attention to those textured regions with high discriminability. As a consequence, the proposed RFM reaches a significant improvement compared with the previous state-of-the-art methods in two publicly available datasets RegDB and SYSU-MM01.

\section*{Acknowledgments}
This work was supported by the National Science Fund for Distinguished Young Scholars (No.62025603), the National Natural Science Foundation of China (No.U1705262, No. 62176222, No. 62176223, No. 62176226, No. 62072386, No. 62072387, No. 62072389, No. 62002305, No. 61772443, No. 61802324 and No. 61702136), Guangdong Basic and Applied Basic Research Foundation (No.2019B1515120049), the Natural Science Foundation of Fujian Province of China (No.2021J01002), and the Fundamental Research Funds for the central universities (No. 20720200077, No. 20720200090 and No. 20720200091).


\bibliographystyle{IEEEtran}
\bibliography{IEEEfull}

\newpage

 




\vfill

\end{document}